\documentclass[sigconf]{acmart}
\AtBeginDocument{%
  }

\setcopyright{acmlicensed}
\copyrightyear{2026}
\acmYear{2026}
\acmDOI{XXXXXXX.XXXXXXX}
\acmConference[KDD'26]{the 32nd ACM SIGKDD Conference on Knowledge Discovery and Data Mining}{August 9--13, 2026}{Jeju, Korea}

\begin{document}


\title{End-to-End Learning for Partially-Observed Time Series with PyPOTS}




\author{Wenjie Du}
\affiliation{
 \institution{\href{https://pypots.com}{\raisebox{-3px}{\includegraphics[width=12px]{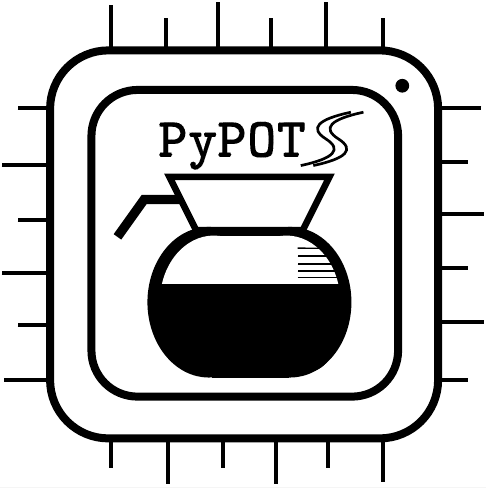}}\hspace{1mm}PyPOTS Research}}
 \city{}
 \country{}
 }
\email{wdu@time-series.ai}

\author{Yiyuan Yang}
\affiliation{
 \institution{PyPOTS Research \& Univ. of Oxford}
 \city{}
 \country{}
}
\email{yiyuan.yang@cs.ox.ac.uk}

\author{Tianxiang Zhan}
\affiliation{%
  \institution{PyPOTS Research}
  \city{}
  \country{}}
\email{ztxtech@foxmail.com}

\author{Qingsong Wen}
\affiliation{%
  \institution{Squirrel Ai Learning}
  \city{}
  \country{}}
\email{qingsongedu@gmail.com}

\renewcommand{\shortauthors}{Du et al.}

\begin{abstract}
Partially-observed time series (POTS) is ubiquitous in real-world applications, yet most existing toolchains separate missing-value handling from downstream learning, which limits reproducibility and overall performance. This tutorial introduces PyPOTS, an open-source Python ecosystem for end-to-end data mining and machine learning on POTS. We present practical workflows spanning missingness simulation, data preprocessing, model training, and evaluation across core tasks, including imputation, forecasting, classification, clustering, and anomaly detection. 

The tutorial consists of two parts: Part I emphasizes hands-on application for practitioners through unified APIs and benchmark-oriented experiments. Part II targets developers and researchers, focusing on extending PyPOTS with custom models, domain-specific constraints, and contribution-ready engineering practices. Participants will gain both conceptual understanding and implementation experience for building robust, transparent, and reusable POTS pipelines in research and production settings. PyPOTS is publicly available at \url{https://github.com/WenjieDu/PyPOTS}.
\end{abstract}

\maketitle

\section{Introduction}

As data-driven systems increasingly rely on multivariate time series from IoT devices, healthcare monitors, industrial sensors, traffic systems, and financial markets, the assumption of fully observed data has become critically flawed~\cite{zhang2025missing,adhikari2022comprehensive,kazijevs2023deep,zhang2024comprehensive,bryzgalova2025missing}. Real-world time series are inherently partially observed due to sensor failures, asynchronous recordings, and communication loss~\cite{10.24963/ijcai.2025/1187,middlehurst2024aeon}. Dealing with incomplete data often leads to fragmented workflows where imputation and downstream predictive tasks are entirely decoupled, resulting in suboptimal performance, limited reusability, and error propagation~\cite{du2024tsi}. Despite the critical need for unified frameworks, most existing libraries offer limited methodologies focused solely on data repair~\cite{nater2025hands,moritz2017imputets,khayati2020mind}, failing to support a comprehensive machine learning lifecycle for partially-observed time series (POTS).

To bridge this gap, we present PyPOTS, a unified Python ecosystem specifically designed for data mining and machine learning on POTS~\cite{du2023pypots}. PyPOTS provides a seamless workflow integrating missingness simulation, preprocessing, state-of-the-art modeling, and evaluation across five core tasks: imputation, forecasting, classification, clustering, and anomaly detection.

\textbf{Learning Outcomes.} The interactive part of the tutorial will provide participants with experience in applying POTS modeling techniques to real-world datasets. Participants will learn how to (1) deploy PyPOTS to build analysis pipelines for time series with varying missingness mechanisms, (2) benchmark diverse state-of-the-art models using unified APIs, (3) execute different end-to-end learning tasks directly on POTS, and (4) extend the PyPOTS ecosystem by integrating and testing custom models or modules. We expect attendees to gain a deep understanding of the POTS paradigm and practical implementation skills.

\textbf{Target Audience and Prerequisites.} Our tutorial is intended for industry practitioners, data scientists, and software engineers dealing with noisy, incomplete real-world data. It also serves researchers focusing on time series analysis and representation learning. Participants should have a basic understanding of Python, Jupyter Notebooks, and fundamental machine learning concepts. Familiarity with PyTorch is helpful for the developer-extension part, but not strictly required for the core engineering workflow.

\textbf{Engaging Experience.} Attendees will benefit from in-depth demonstrations and carefully designed step-by-step hands-on materials. Utilizing Google Colab and pre-configured Jupyter Notebooks, participants will actively engage in live coding exercises, pipeline construction, and model benchmarking, ensuring a smooth, scalable, and installation-free experience during the tutorial.

\section{Tutorial Outline}
This \textbf{system-focused} tutorial is organized into two tightly connected parts for both practitioners and developers. Part I emphasizes rapid, end-to-end application with reproducible notebooks, while Part II focuses on architecture-level extension and contribution-ready development practices. The session follows a progressive pattern of concept briefing, live coding, guided exercises, and short discussion checkpoints. We recommend approximately 60\% of the time for Part I and 40\% for Part II, with brief transitions to connect practical usage and internal extension design.

\textit{\textbf{Part I: Apply PyPOTS to Time Series Analysis:}}
Part I focuses on hands-on practice for participants who want to build reliable POTS pipelines with minimal engineering overhead quickly: 
\textbf{I.1 Problem setup and POTS fundamentals.} We introduce common missingness mechanisms, discuss why complete-data assumptions fail in real applications, and define task-oriented evaluation principles for partially observed time series.
\textbf{I.2 Data preparation and missingness simulation.} Using PyPOTS utilities, we demonstrate dataset loading, preprocessing, train/validation/test splitting, and controlled missingness injection for reproducible benchmarking.
\textbf{I.3 Unified model training for time series analysis tasks.} Participants run representative PyPOTS models with a consistent API, compare reconstruction quality under different missing rates, and analyze robustness-efficiency tradeoffs. We show how to perform imputation, forecasting, classification, clustering, and anomaly detection under partial observation.
\textbf{I.4 Evaluation, visualization, and reproducibility checklist.} We provide practical guidance on metric selection, error analysis, visualization, and experiment logging to produce transparent and reproducible results.
\textbf{Hands-on outcome.} By the end of Part I, attendees complete an end-to-end notebook pipeline from raw incomplete data to model comparison and report-ready evaluation artifacts.

\textit{\textbf{Part II: Extend PyPOTS to Specialties:}}
Part II targets developers and researchers who plan to customize PyPOTS for domain-specific requirements and novel methods: 
\textbf{II.1 PyPOTS architecture walkthrough.} We explain the modular design of datasets, models, training loops, and evaluation components, highlighting extension points for new algorithms and tasks.
\textbf{II.2 Implementing a custom model/module.} Starting from a minimal template, we demonstrate how to add a new method (or backbone block), define forward and loss logic, and connect it to the training interface.
\textbf{II.3 Supporting domain constraints.} We discuss practical customization patterns, including irregular sampling handling, feature-wise masking strategies, and task-specific objective design for healthcare, IoT, and industrial scenarios.
\textbf{II.4 Benchmark integration and testing.} Participants learn how to plug custom modules into existing benchmarking workflows, run sanity checks, and ensure fair comparisons against built-in baselines.
\textbf{II.5 Open-source workflow and contribution practice.} We cover coding conventions, documentation expectations, test integration, and pull-request best practices to help attendees contribute extensions back to the PyPOTS ecosystem.
\textbf{Hands-on outcome.} By the end of Part II, attendees produce a runnable extension prototype with reproducible experiment scripts and a contribution-ready project structure.

\enlargethispage{\baselineskip}
\section{Societal Impact}
POTS is pervasive in domains such as healthcare, IoT, industrial monitoring, etc. By helping researchers and developers build more realistic workflows, this tutorial contributes to more effective modeling of POTS and more reliable use in real-world scenarios. Crucially, by promoting standardized building pipelines and evaluation protocols, the tutorial aims to significantly improve research reproducibility across the time series community. At the same time, outputs derived from POTS should not be treated as ground truth and are intended for reference.

\enlargethispage{\baselineskip}
\section{Related Materials}

Slides, executable notebooks, and code examples will be released through the official PyPOTS portal (\url{https://pypots.com/}), documentation site (\url{https://docs.pypots.com/}), the PyPOTS GitHub repository (\url{https://github.com/WenjieDu/PyPOTS/}), and the BrewPOTS tutorial repository (\url{https://github.com/WenjieDu/BrewPOTS/}).

\enlargethispage{\baselineskip}
\section{Tutors’ Biography}
Wenjie Du is the founder of PyPOTS Research. He holds an M.A.Sc. degree and focuses on building AI systems for time series analysis and real-world applications.

Yiyuan Yang is a PhD student at the University of Oxford. His research focuses on intelligent sensing, time series, and spatiotemporal data mining. He previously studied at Tsinghua University and interned at Microsoft, Alibaba, and Huawei.

Tianxiang Zhan is a researcher at PyPOTS Research. His interests include time series analysis, information theory, and complex systems.

Qingsong Wen is the Head of AI and Chief Scientist at Squirrel Ai Learning. He holds a PhD from Georgia Tech and has published 200+ papers in top venues (e.g., KDD, NeurIPS, ICML, ICLR). He chairs the IEEE CIS Task Force on AI for Time Series and serves as (Senior) Associate Editor for IEEE TPAMI and IEEE TSP. He also organized multiple time series workshops and tutorials at KDD.

\bibliographystyle{ACM-Reference-Format}
\bibliography{sample-base}

\end{document}